\documentclass[runningheads]{llncs}
\usepackage[centering]{geometry}

\usepackage{amsfonts}
\usepackage{amsmath}
\usepackage[applemac]{inputenc}
\usepackage[english]{babel}
\usepackage{color}
\usepackage{hyperref}
\usepackage{subcaption}

\usepackage{amssymb}
\usepackage{amsopn}
\usepackage{mathrsfs}

\usepackage{mathtools}
\mathtoolsset{showonlyrefs}

\DeclareMathOperator{\supp}{supp}

\newcommand{\cE}{\mathcal{E}}
\newcommand{\cI}{\mathcal{I}}

\newcommand{\bT}{\mathbb{T}}

\newcommand{\cU}{\mathcal{U}}

\newcommand{\R}{\mathbb{R}}

\newcommand{\be}{\begin{equation}}
\newcommand{\ee}{\end{equation}}

\newcommand{\bS}{\mathbb{S}}

\DeclareMathOperator{\bR}{\mathbb{R}}

\DeclareMathOperator{\rng}{rng}

\newtheorem{prop}[theorem]{Proposition}

\usepackage[normalem]{ulem}

\usepackage{todonotes}
\allowdisplaybreaks

\begin{document}
\title{A cortical-inspired model for orientation-dependent contrast perception: a link with Wilson-Cowan equations.} 
\titlerunning{Cortical-inspired Wilson-Cowan modelling for contrast perception}
%
 \author{Marcelo Bertalm\'io\inst{1}
  \and
 Luca Calatroni\inst{2}
  \and
 Valentina Franceschi\inst{3}
  \and \\
 Benedetta Franceschiello\inst{4}
 \and
 Dario Prandi\inst{5}
 }
 \authorrunning{Bertalm\'io et al.}
 %
 \institute{DTIC, Universitat Pompeu Fabra, Barcelona, Spain (\email{marcelo.bertalmio@upf.edu}) \and
 CMAP, \'Ecole Polytechnique CNRS, Palaiseau, France
 (\email{luca.calatroni@polytechnique.edu}) \and
 IMO, Universit\'e Paris-Sud, Orsay, France (\email{valentina.franceschi88@gmail.com}) \and
 Fondation Asile des Aveugles (FAA) and Laboratory for Investigative Neurophysiology (LINE), Radiology, CHUV, Lausanne, Switzerland (\email{benedetta.franceschiello@gmail.com}) \and 
 CNRS, L2S, Centrale Supelec, France (\email{dario.prandi@l2s.centralesupelec.fr})}
\maketitle              
\begin{abstract}
We consider a differential model describing neuro-physiological contrast perception phenomena induced by surrounding orientations. The mathematical formulation relies on a cortical-inspired modelling  \cite{Citti2006} largely used over the last years to describe neuron interactions in the primary visual cortex (V1) and applied to several image processing problems \cite{Duits2010,Prandi2017,Franceschiello2018}. Our model connects to Wilson-Cowan-type equations \cite{WilsonCowan1973} and it is analogous to the one used in \cite{Bertalmio2007,BertalmioFrontiers2014,KimBatardBertalmio2016} to describe assimilation and contrast phenomena, the main novelty being its explicit dependence on local image orientation. To confirm the validity of the model, we report some numerical tests showing its ability to explain orientation-dependent phenomena (such as grating induction) and geometric-optical illusions \cite{Weintraub1971,McCourt1982} classically explained only by filtering-based techniques \cite{Blakeslee1999,Otazu2008}.

\end{abstract}

\section{Introduction}  \label{sec:intro}

Many, if not most, popular vision models consist in a cascade of linear and non-linear (L+NL) operations \cite{Martinez2018}. This happens for models describing both visual perception -- e.g., the Oriented  Difference  Of  Gaussians  (ODOG)  \cite{Blakeslee1999} or the Brightness  Induction  Wavelet  Model (BIWaM) \cite{Otazu2008} -- and  neural activity  \cite{Carandini2005}.
However, L+NL models, while suitable in many cases for retinal and thalamic activity, are not adequate to predict neural activity in the primary visual cortex area (V1). In fact, according to \cite{Carandini2005}, such models have low predictive power (indeed they can explain less than 40\% of the variance of the data).
On the other hand, several vision models are not in the form of a cascade of L+NL operations, such as those describing neural dynamics via Wilson-Cowan (WC) equations \cite{WilsonCowan1973,Bressloff2002}.
These equations describe the state $a(x,\theta,t)$ of a population of neurons with V1 coordinates $x\in\mathbb R^2$ and orientation preference $\theta\in [0,\pi)$ at time $t>0$ as
\begin{equation}\label{eq:WC}
    \frac{\partial}{\partial t} a(x,\theta,t) = -\alpha a(x,\theta,t) + \nu \int_{0}^\pi\int_{\mathbb R^2} \omega(x,\theta \| x', \theta') \sigma(a(x',\theta',t))\,dx'\,d\theta' + h(x,\theta,t).
\end{equation}
Here, $\alpha,\nu>0$ are fixed parameters, $\omega(x,\theta\|x',\theta')$ is an interaction weight, $\sigma:\mathbb R\to\mathbb R$ is a sigmoid saturation function and $h$ represents the external stimulus. In \cite{Bertalmio2007,BertalmioCowan2009,BertalmioFrontiers2014} the authors show how orientation independent WC-type equations admit a variational formulation through an associated energy functional which can be linked to histogram equalisation, visual adaptation and efficient coding theories
\cite{Olshausen2000}.

 In this paper, we consider a generalisation of this modelling and introduce explicit orientation dependence via a lifting procedure inspired by neuro-physiological models of V1 \cite{Citti2006,Duits2010,Prandi2017}. Interestingly, the Euler-Lagrange equations associated with the proposed functional yield orientation-dependent WC-type equations analogous to \eqref{eq:WC}. We then report some numerical evidence showing how the proposed model is able to better reproduce some visual perception phenomena in comparison to both previous orientation-independent WC-type models and  state-of-the-art ODOG \cite{Blakeslee1999} and BIWaM \cite{Otazu2008} L+NL models.
%
%
%
%

In particular, we firstly test our model on orientation-dependent Grating Induction (GI) phenomena (generalising the ones presented in \cite[Figure~3]{Blakeslee1999}, see also \cite{McCourt1982}) and show a direct dependence of the processed image on the orientation, which cannot be reproduced via orientation-independent models, but which is in accordance with the reference result provided by the ODOG model. We then test the proposed model on the Poggendorff illusion, a geometrical optical effect where a misalignment of two collinear segment is induced by the presence of a surface \cite{Weintraub1971,Westheimer2008}, see Figure \ref{fig:poggendorff}. For this example our model is able to reconstruct the perceptual bias better than all the reference models considered.

\section{Orientation-independent variational models} \label{sec:review}


Let $Q\subset\R^2$ be a rectangular image domain with and let $f:Q\to[0,1]$ be a normalised image on $Q$. Let further $\omega:Q\times Q\to \bR_+$ be a given positive symmetric weighting function such that for any $x\in Q,~ \omega(x,\cdot)\in L^1(Q)$. For a fixed parameter $\alpha>1$ we consider the piece-wise affine sigmoid $\sigma_\alpha:\mathbb R\to[-1,1]$ defined by
\begin{equation}  \label{def:sigmoid}
  \sigma_\alpha(\rho) := \min\{1,\max\{\alpha \rho, -1\}\},
\end{equation}
and we consider $\Sigma_\alpha$ to be any function such that $\Sigma_\alpha'=\sigma_\alpha$. Observe that $\Sigma_\alpha$ is convex and even.

In \cite{Bertalmio2007,BertalmioFrontiers2014,KimBatardBertalmio2016} the authors consider the following energy as a variational model for modelling contrast and assimilation phenomena, defined in terms of a given initial image $f_0$:
\begin{multline}
  \cE(f):= \frac12 \int_{Q}\left(f(x)-\mu(x)\right)^2dx + \frac{\lambda}{2}\int_{Q}\left(f(x)-f_0(x)\right)^2dx \\
  - \frac{1}{4M}\int_{Q}\int_{Q} \omega(x,y)\Sigma_\alpha\big( f(x)-f(y) \big)\,dx\,dy. \label{eq:energy-bertalmio}
\end{multline}
Here, $\mu:Q\to \mathbb R$ is the local mean of the initial image $f_0$.
Such term can encode a global reference to the ``Grey-World'' principle \cite{Bertalmio2007} (in this case $\mu(x)=\frac12$ for any $x\in Q$) or a filtering around $x\in Q$ computed either via a single-Gaussian convolution \cite{BertalmioFrontiers2014} or  a sum of Gaussian filters \cite{KimBatardBertalmio2016}, consistently with the modelling of the multiple inhibition effects happening at a retinal-level \cite{MarceloPLOS2016}.
Finally, the parameter $M\in (0,1]$ stands for a normalisation constant, while $\lambda>0$ represents a weighting parameter enforcing the attachment of the solution $f$ to the given $f_0$.

The gradient descent associated with $\cE$ is the following equation corresponds to a Wilson-Cowan-type  equation similar to \eqref{eq:WC}, where the visual activation is assumed to be independent of the orientation, as discussed in \cite{Bressloff2001}. The parameters $\alpha$ and $\nu$ are set $\alpha=1$ and $\nu=\frac{1}{2M}$ and the time-invariant external stimulus $h$ is equal to $\mu + \lambda f_0$:
\begin{equation}  \label{eq:PDE_bertalmio}
  \frac{\partial}{\partial t} f(x,t)
  = -(1+\lambda)f(x,t)+\left(\mu(x)+ \lambda f_0(x)\right)+ \frac{1}{2M}\int_{\bT^2} \omega(x,y)\sigma_\alpha\big( f(x,t)-f(y,t) \big)\,dy.  
\end{equation}

Note that in \cite{KimBatardBertalmio2016} the authors consider in  \eqref{eq:energy-bertalmio} a convolution kernel $\omega$ which is a convex combination of two bi-dimensional Gaussians with different standard deviations. While this variation of the model is effective in describing \emph{assimilation} effects, the lack of dependence on the local perceived orientation in the Wilson-Cowan-type equation \eqref{eq:PDE_bertalmio} makes such modelling intrinsically not adapted to explain orientation-induced contrast and colour perception effects similar to the ones observed in \cite{Otazu2008,Self2014,Blakeslee1999}. Up to our knowledge, the only perceptual models capable to explain these effects are the ones based on oriented Difference of Gaussian filtering coupled with some non-linear processing, such as the ODOG and the BIWaM models described in \cite{Blakeslee1999,Blakeslee2016} and  \cite{Otazu2008}, respectively.


\section{A cortical-inspired model}

Let us denote by $R>0$ the size of the visual plane and let $D_R\subset \bR^2$ be the disk $D_R:=\{x_1^2+x_2^2 \le R^2\}$. Fix $R>0$ such that $Q\subset D_R$.
In order to exploit the properties of the roto-translation group $SE(2)$ on images, we  now consider them to be elements in the set:
$$
  \cI = \left\{f \in L^2(\bR^2,[0,1]) \text{ such that\ } \supp f\subset D_R\right\}.
$$
We remark that fixing $R>0$ is necessary, since contrast perception is strongly dependent on the scale of the features under consideration w.r.t.\ the visual plane. 

In the following, after introducing the functional lifting under consideration, which follows well-established ideas contained, e.g., in \cite{Citti2006,Duits2010,Prandi2017}, we present the proposed cortical-inspired extension of the energy functional \eqref{eq:energy-bertalmio}, and discuss the numerical implementation of the associated gradient descent. We remark that the resulting procedure thus consists of a linear lifting step combined with WC-type evolution, which is non-linear due to the presence of the sigmoid $\sigma_\alpha$, see \eqref{def:sigmoid}.

\subsection{Functional lifting} \label{sec:lifting}

Each neuron $\xi$ in V1 is assumed to be associated with a receptive field (RF) $\psi_\xi\in L^2(\bR^2)$ such that its response under a visual stimulus $f\in \mathcal I$ is given by 
\begin{equation}\label{eq:rf}
    F(\xi) = \langle \psi_\xi, f\rangle_{L^2(\bR^2)} = \int_{\bR^2} \overline{\psi_\xi(x)} f(x)\, dx.
\end{equation}
Motivated by neuro-phyisiological evidence, we will assume that each neuron is sensible to a preferred position and orientation in the visual plane, i.e., that $\xi=(x,\theta)\in \mathcal{M} = \mathbb R^2\times \mathbb P^1$. Here, $\mathbb P^1$ is the projective line that we represent as $[0,\pi]/\sim$, with $0\sim \pi$. Moreover, in order to respect the \emph{shift-twist} symmetry \cite[Section ~4]{Bressloff2002}, we will assume that the RF of different neurons are ``deducible'' one from the other via a linear transformation. Let us explain this in detail.

The double covering of $\mathcal{M}$ is given by the Euclidean motion group $SE(2)=\bR^2\rtimes \bS^1$, that we consider endowed with its natural semi-direct product structure
$$
  (x,\theta)\star(y,\varphi) 
  = 
 (x+R_\theta y, \theta+\varphi), 
 \qquad
 \forall (x,\theta),(y,\varphi)\in SE(2), 
 \qquad
 R_\theta = \left(\begin{array}{cc}
     \cos\theta & -\sin\theta \\
      \sin\theta & \cos\theta
 \end{array}\right).
$$
In particular, the above operation induces an action of $SE(2)$ on $\mathcal{M}$, which is thus an homogeneous space. 
Observe that $SE(2)$ is unimodular and that its Haar measure (which is the only left and right-invariant measure up to scalar multiples) is simply $dxd\theta$.

We now denote by $\cU(L^2(\bR^2)) \subset \mathcal{L}(L^2(\bR^2))$ the space of linear unitary operators on $L^2(\bR^2)$ and let $\pi:SE(2)\to \cU(L^2(\bR^2))$ be the \emph{quasi-regular representation} of $SE(2)$ which associates to any $(x,\theta)\in SE(2)$ the unitary operator $\pi(x,\theta)\in \cU(L^2(\bR^2))$, i.e., the action of the roto-translation $(x,\theta)$ on square-integrable functions on $\mathbb R^2$. Namely, the operator $\pi(x,\theta)$ acts on $\psi\in L^2(\bR^2)$ by
\begin{equation*}
  [\pi(x,\theta)\psi](y) = \psi((x,\theta)^{-1}y)  = \psi(R_{-\theta}(y-x)), \qquad \forall y\in \R^2.
\end{equation*}
Moreover, we let $\Lambda:SE(2)\to \cU(L^2(SE(2)))$ be the \emph{left-regular representation}, which acts on functions $F\in L^2(SE(2))$ as
\begin{equation*}
  [\Lambda(x,\theta)F](y,\varphi) = F((x,\theta)^{-1}\star (y,\varphi)) = F(R_{-\theta}(y-x), \varphi-\theta), \quad \forall (y,\theta)\in SE(2).
\end{equation*}

Letting now $L:L^2(\bR^2)\to L^2(\mathcal{M})$ be the operator that transforms visual stimuli into cortical activations, one can formalise the \emph{shift-twist} symmetry by requiring that
\begin{equation}
    L\circ \pi(x,\theta) = \Lambda(x,\theta)\circ L, \qquad \forall (x,\theta)\in SE(2).
\end{equation}
Under mild continuity assumption on $L$,  it has been shown in \cite{Prandi2017} that $L$ is then a continuous wavelet transform. That is, there exists a \emph{mother wavelet} $\Psi\in L^2(\bR^2)$ satisfying $\pi(x,\theta)\Psi = \pi(x,\theta+\pi)\Psi$ for all $(x,\theta)\in SE(2)$, and  such that
\begin{equation}\label{eq:wavelet}
Lf(x,\theta) = \langle \pi(x,\theta)\Psi, f \rangle, \qquad \forall f\in L^2(\bR^2), (x,\theta)\in \mathcal{M}.
\end{equation}
Observe that the operation $\pi(x,\theta)\Psi$ above is well defined for $(x,\theta)\in \mathcal{M}$ thanks to the assumption on $\Psi$.
By \eqref{eq:rf}, the above representation of $L$ is equivalent to the fact that the RF associated with the neuron $(x,\theta)\in \mathcal{M}$ is the roto-translation of the mother wavelet, i.e., $\psi_{(x,\theta)}=\pi(x,\theta)\Psi$.

\begin{remark}
  Letting $\Psi^*(x):=\overline{\Psi(-x)}$, the above formula can be rewritten as
\begin{equation}
    Lf(x,\theta) 
      = \int_{\bR^2} \overline{\Psi(R_{-\theta}(y-x))}f(y)\,dy  
      = \big[f * (\Psi^*\circ R_{-\theta})\big] (x),
\end{equation}
where $f*g$ denotes the standard convolution on $L^2(\bR^2)$.
\end{remark}

Notice that, although images are functions of $L^2(\bR^2)$ with values in $[0,1]$, it is in general not true that $Lf(x,\theta)\in [0,1]$. 
However, from \eqref{eq:wavelet} we deduce the following result, which guarantees that $L$ is a bounded operator. (See \cite[Section~2.2.3]{Prandi2017}.)

\begin{prop}\label{prop:max-lift}
  The operator $L:L^2(\bR^2)\to L^2(\mathcal{M})$ is continuous and, therefore, bounded.
  In particular, for any $f\in L^2(\bR^2)$ the function $Lf$ is a continuous bounded function on $M$, with $|Lf|\le \|\Psi\|_{L^2(\bR^2)}\|f\|_{L^2(\bR^2)}$.
  Moreover, if $f$ takes values in $[0,1]$ and $\Psi\in L^1(\bR^2)$, we have $|Lf|\le \|\Psi\|_{L^1(\bR^2)}$.
\end{prop}

Neuro-physiological evidence shows that a good fit for the RFs is given by Gabor filters, whose Fourier transform is simply the product of a Gaussian with an oriented plane wave \cite{Daugman1985a}.  However, these filters are quite challenging to invert, and are parametrised on a bigger space than $\mathcal M$, which takes into account also the frequency of the plane wave and not only its orientation. For this reason, in this work we chose to consider as wavelets the \emph{cake wavelets} introduced in \cite{Bekkers2014}. These are obtained via a mother wavelet $\Psi^{\text{cake}}$ whose support in the Fourier domain is concentrated on a fixed slice, which depends on the number of orientations one aims to consider in the numerical implementation. To recover integrability properties, the Fourier transform of this mother wavelet is then smoothly cut off via a low-pass filtering, see \cite[Section ~2.3]{Bekkers2014} for details. Observe, however, that in order to lift to $\mathcal M$ and not to $SE(2)$, we consider a non-oriented version of the mother wavelet, given by $\tilde\psi^{cake}(\mathbf{\omega}) + \tilde\psi^{cake}(e^{i\pi}\mathbf{\omega})$, in the notations of \cite{Bekkers2014}.

An important feature of cake wavelets is that, in order to recover the original image, it suffices to consider the \emph{projection operator} defined by
\begin{equation}\label{eq:proj}
    P: L^2(\mathcal M)\to L^2(\bR^2), \qquad
    PF(x) := \int_{\mathbb P^1} F(x,\theta)\,d\theta,\qquad F\in L^2(\mathcal{ M})
\end{equation}
Indeed, by construction of cake wavelets, Fubini's Theorem shows that $(P\circ L)f = f$ for all $f\in \mathcal I$.


\subsection{WC-type mean field modelling}     \label{sec:cortical_model}

The natural extension of \eqref{eq:energy-bertalmio}  to the cortical setting introduced in the previous section is the following energy on functions $F\in L^2(\mathcal{M})$:
\begin{multline}   \label{eq:tentative_energy}
  \cE(F) = \frac12 \left\|F-G_0\right\|^2_{L^2(\mathcal{M})} + \frac{\lambda}{2} \left\|F-F_0\right\|^2_{L^2(\mathcal{M})}\\
  - \frac{1}{4M}\int_{\mathcal{M}}\int_{\mathcal{M}} \omega(x,\theta\|x',\theta')\Sigma_\alpha\big( F(x,\theta)-F(x',\theta') \big)\,dxd\theta\,dx'd\theta'.
\end{multline}
Here, $G_0$ is the lift $L\mu$ of the local mean appearing in \eqref{eq:energy-bertalmio}, and $F_0$ is the lift $Lf_0$ of the given initial image $f_0$. 
The constant $M\in (0,1]$ will be specified later.
Furthermore, $\omega:\mathcal{M}\times \mathcal{M}\to \bR_+$ is a positive symmetric weight function  and, as in \cite{Bertalmio2007}, we denote by $\omega(x,\theta\|x',\theta')$ its evaluation at $((x,\theta),(x',\theta'))$.  A sensible choice for $\omega$ would be an approximation of the heat kernel of the anisotropic diffusion associated with the structure of V1, as studied in \cite{Citti2006,Duits2010}.
However, in this work we chose to restrict to the case where $\omega(x,\theta\| x', \theta')=\omega(x-x', \theta-\theta')$ is a (normalised) three-dimensional Gaussian in $\mathcal{M}$, since this is sufficient to describe many perceptual phenomena not explained by the previous 2D model \eqref{eq:energy-bertalmio}. 
A possible improvement could be obtained by choosing $\omega$ as the sum of two such Gaussians at different scales in order to better describe, e.g., lightness assimilation phenomena.

Thanks to the symmetry of $\omega$ and the oddness of $\sigma$, computations similar to those of \cite{Bertalmio2007} yield that the gradient descent associated with \eqref{eq:tentative_energy} is
\begin{multline}\label{eq:proj-wc}
      \frac{\partial}{\partial t} F (x,\theta,t)
      = -(1+\lambda)F(x,\theta,t)+G_0(x,\theta)+\lambda F_0(x,\theta) \\
        +\frac1{2M} \int_{\mathcal M} \omega(x,\theta \|x',\theta')\sigma_\alpha\big( F(x,\theta,t)-F(x',\theta',t)\big)\,dx'd\theta'.
\end{multline}
This is indeed a Wilson-Cowan-like equation, where the external cortical stimulus is $h = G_0+\lambda F_0$.

We observe that in general there is no guarantee that $\rng L$ is invariant w.r.t.\ the evolution $t\mapsto F(\cdot,\cdot,t)$ given by \eqref{eq:proj-wc}. That is, in general, although $F_0=Lf_0$, if $t>0$ there is no image $f(\cdot,t)\in L^2(\bR^2)$ such that $F(\cdot,\cdot,t)=Lf(\cdot,\cdot,t)$. 
We will nevertheless assume that the perceived image for any cortical activation is given by the projection operator \eqref{eq:proj}.


\subsection{Discretisation via gradient descent}\label{sec:num_impl}

In order to numerically implement the gradient descent \eqref{eq:proj-wc}, we discretise the initial (square) image $f_0$ as an $N\times N$ matrix. (Here, we assume periodic boundary conditions.) We additionally consider $K\in \mathbb N$  orientations, parametrised by $k \in \{1,\ldots,K\}\mapsto \theta_k := (k-1)\pi/K$. 

The discretised lift operator, still denoted by $L$, then transforms $N\times N$ matrices into $N\times N\times K$ arrays. Its action on an $N\times N$ matrix $f$ is defined by
\begin{equation}
  (Lf)_{n,m,k} = \mathcal F^{-1}\left( (\mathcal F f) \odot (R_{\theta_k} \mathcal F\Psi^{\text{cake}}) \right)_{n,m}
  \qquad \forall n,m\in \{1,\ldots, N\}, \, k\in \{1,\ldots,K\},
\end{equation}
where $\odot$ is the Hadamard (i.e., element-wise) product of matrices, $\mathcal F$ denotes the discrete Fourier transform, $R_{\theta_k}$ is the rotation of angle $\theta_k$, and  $\Psi^{\text{cake}}$ is the cake mother wavelet.

After denoting by $F^0 = Lf_0$, and by $G_0 = L\mu$ where $\mu$ is a Gaussian filtering of $f_0$, we find that the explicit time-discretisation of the the gradient descent \eqref{eq:proj-wc} is
\begin{equation}
    \frac{F^{\ell+1} - F^\ell}{\Delta t}
    = 
    -(1+\lambda)F^\ell + G_0 +\lambda F^0 + \frac{1}{2M} R_{F^\ell}, \qquad \Delta t\ll 1, \, \ell\in\mathbb N.
\end{equation}
Here, for a given 3D Gaussian matrix $W$  encoding the weight $\omega$, and an $N\times N\times M$ matrix $F$, we let, for any $n,m\in \{1,\ldots, N\}$ and $k\in\{1,\ldots, K\}$, 
\begin{equation}
    (R_{F})_{n,m,k} := \sum_{n',m'=1}^N\sum_{k'=1}^K W_{n-n', m-m', k-k'} \sigma( F_{n,m,k} - F_{n',m',k'}  ).
\end{equation}
We refer to \cite[Section ~IV.A]{Bertalmio2007} for the description of an efficient numerical approach  used to compute the above quantity in the 2D case and that can be translated verbatim to the 3D case under consideration.

After a suitable number of iterations $ \bar \ell$ of the above algorithm (measured by the criterion $\|F^{\ell+1}-F^\ell\|_2/\|F^\ell\|_2\le \tau$, for a fixed tolerance $\tau\ll 1$), the output image is then found via \eqref{eq:proj} as
\begin{equation}
    \bar f_{n,m} = \sum_{k=1}^K F^{\bar \ell}_{n,m,k}.
\end{equation}

\section{Numerical results}  \label{sec:numres}

\begin{figure}[t]
\centering
\begin{subfigure}{0.45\textwidth}
    \centering
   \includegraphics[height=4cm]{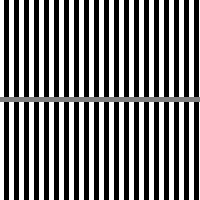}
   \caption{Relative orientation $\theta=\pi/2$.}
   \label{fig:barrette1}
\end{subfigure}
\begin{subfigure}{0.45\textwidth}
\centering
    \includegraphics[height=4cm]{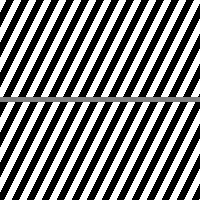}
    \caption{Relative orientation $\theta=\pi/3$.}
     \label{fig:barrette2}
 \end{subfigure}
    \caption{Grating inductions with varying background orientation used in the experiments. }
    \label{fig:barrette_mult_or}
\end{figure}

In this section we present results obtained by applying the cortical-inspired model presented in the previous section to a class of well-established phenomena where contrast perception is affected by local orientations. 

We compare the results obtained by our model with the corresponding Wilson-Cowan-type 2D model \eqref{eq:energy-bertalmio}-\eqref{eq:PDE_bertalmio} for contrast enhancement considered in \cite{KimBatardBertalmio2016,BertalmioFrontiers2014}. For further reference, we also report comparisons with two standard reference models based on oriented Gaussian filtering. The former is the ODOG model \cite{Blakeslee1999} where the output is computed via a convolution of the input image with oriented difference of Gaussian filters in six orientations and seven spatial frequencies. The filtering outputs within the same orientation are then summed in a non-linear fashion privileging higher frequencies. The latter model used for comparison is the BIWaM model, introduced in \cite{Otazu2008}. This is a variation of the ODOG model, the difference being the dependence on the local surround orientation of the contrast sensitivity function\footnote{For our comparisons we used the ODOG and BIWaM codes freely available at \url{https://github.com/TUBvision/betz2015_noise}. }.

\paragraph{Parameters.} All the considered images are $200 \times 200$ pixel. We always consider lifts with $K=30$ orientations. The relevant cake wavelets are then computed following \cite{Bekkers2014} for which the frequency band \texttt{bw} is set to $\texttt{bw}=4$ for all experiments. In \eqref{eq:proj-wc}, we compute the local mean average $\mu$ and the integral term by Gaussian filtering with standard deviation $\sigma_\mu$ and $\sigma_\omega$, respectively. The gradient descent algorithm stops when the relative stopping criterion defined in Section~\ref{sec:num_impl} with a tolerance $\tau = 10^{-2}$ is verified.

\subsection{Grating induction with oriented relative background}  \label{sec:grating}

\begin{figure}[t]
\centering
\begin{subfigure}{0.24\textwidth}
    \centering
   \includegraphics[width=\textwidth]{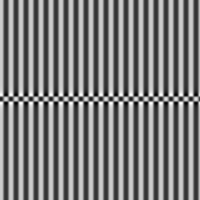}
   \caption{ODOG result.}
   \label{fig:recODOG}
\end{subfigure}
 \begin{subfigure}{0.24\textwidth}
\centering
    \includegraphics[width=\textwidth]{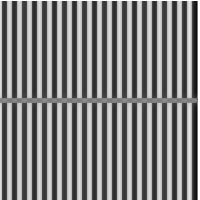}
    \caption{BIWaM result.}
     \label{fig:rec3D}
 \end{subfigure}
\begin{subfigure}{0.24\textwidth}
\centering
    \includegraphics[width=\textwidth]{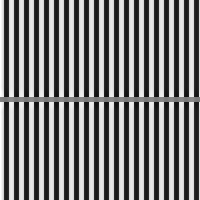}
    \caption{2D WC-type model.}
    \label{fig:rec2D}
 \end{subfigure}
 \begin{subfigure}{0.24\textwidth}
\centering
    \includegraphics[width=\textwidth]{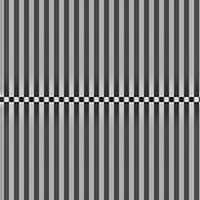}
    \caption{3D WC-type model.}
     \label{fig:rec3D}
 \end{subfigure}
    \caption{Model outputs of input Figure \ref{fig:barrette1}. Parameters for (d): $\sigma_{\mu} = 10$, $\sigma_\omega = 5$, $\lambda = 0.5$.}
    \label{fig:reconstructions_barrette}

\centering
\begin{subfigure}{0.45\textwidth}
    \centering
   \includegraphics[height=4cm]{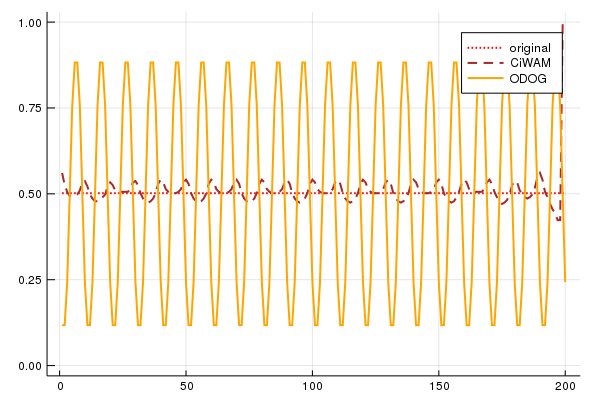}
   \caption{ODOG and BIWaM.}
   \label{fig:lineODOGBIWaM}
\end{subfigure}
\begin{subfigure}{0.45\textwidth}
\centering
    \includegraphics[height=4cm]{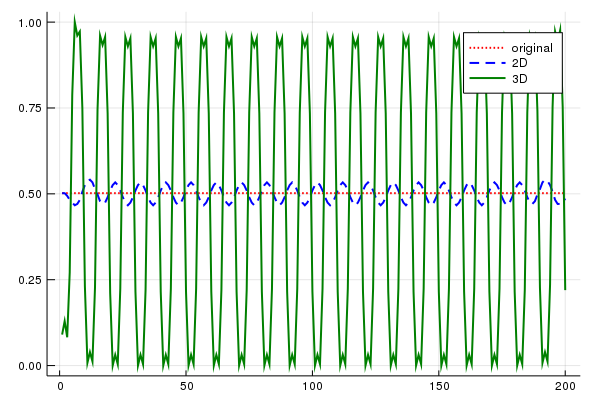}
    \caption{2D VS 3D algorithm.}
    \label{fig:line2D3D}
 \end{subfigure}
    \caption{Middle line-profiles of outputs in Figure \ref{fig:reconstructions_barrette}.  }
    \label{fig:line_profiles_barrette}
\end{figure}

Grating induction (GI) is a contrast effect which has been first described in \cite{McCourt1982} and later studied, e.g., in \cite{Blakeslee1999}. In this section we describe our results about GI with \emph{relative background orientation}, see Figure \ref{fig:barrette_mult_or}. 
Here, when the background has different orientation from the central grey bar, a grating effect, i.e. an alternation of dark-grey/light-grey patterns within the central bar is produced and perceived by the observer.
This phenomenon is contrast dependent, as the intensity of the induced  grey patterns (dark-grey/light-grey) is in opposition with the background grating.
Moreover, it is also orientation-dependent as the magnitude of the phenomenon increase or decrease based on the background orientation, and is maximal when the background bars are orthogonal to the central grey bar.

\begin{figure}[t]
\centering
\begin{subfigure}{0.24\textwidth}
    \centering
   \includegraphics[width=\textwidth]{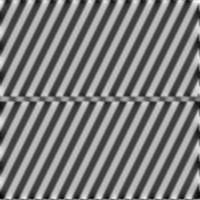}
   \caption{ODOG result.}
   \label{fig:recODOG2}
\end{subfigure}
\begin{subfigure}{0.24\textwidth}
\centering
    \includegraphics[width=\textwidth]{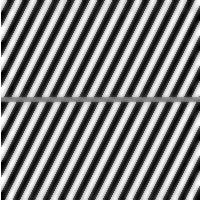}
    \caption{BIWaM result.}
     \label{fig:rec3D}
 \end{subfigure}
\begin{subfigure}{0.24\textwidth}
\centering
    \includegraphics[width=\textwidth]{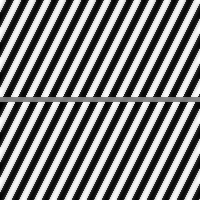}
    \caption{2D WC-type model.}
    \label{fig:rec2D2}
 \end{subfigure}
 \begin{subfigure}{0.24\textwidth}
\centering
    \includegraphics[width=\textwidth]{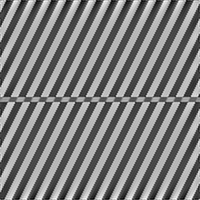}
    \caption{3D WC-type model.}
     \label{fig:rec3D2}
 \end{subfigure}
    \caption{Model outputs of input Figure \ref{fig:barrette2}. Parameters for (d): $\sigma_{\mu} = 10$, $\sigma_\omega = 5$, $\lambda = 0.5$. }
    \label{fig:reconstructions_barrette2}

\centering
\begin{subfigure}{0.45\textwidth}
    \centering
   \includegraphics[height=4cm]{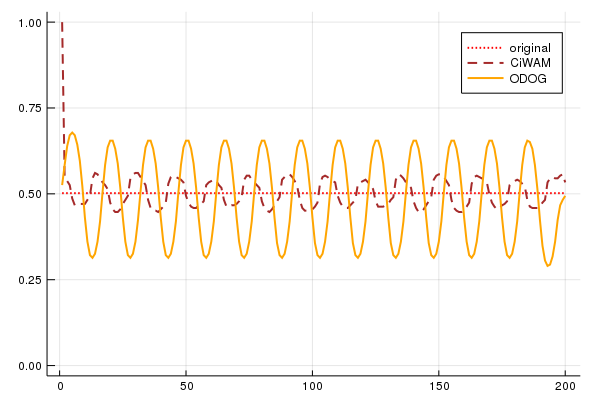}
   \caption{ODOG and BIWaM.}
   \label{fig:lineODOGBIWaM2}
\end{subfigure}
\begin{subfigure}{0.45\textwidth}
\centering
    \includegraphics[height=4cm]{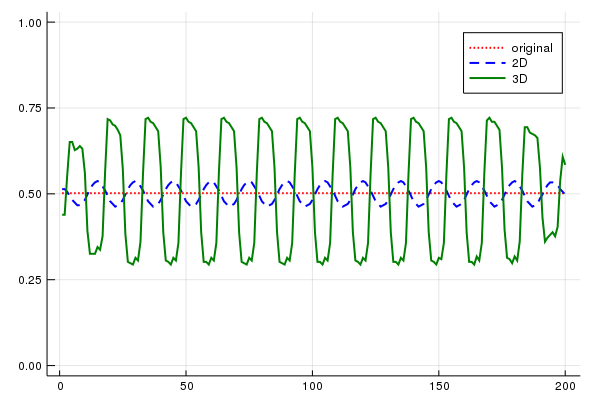}
    \caption{2D VS 3D algorithm.}
    \label{fig:line2D3D2}
 \end{subfigure}
    \caption{Middle line-profiles of outputs in Figure \ref{fig:reconstructions_barrette2}.  }
    \label{fig:line_profiles_barrette2}
\end{figure}

\paragraph{Discussion on computational results.} We observe that model \eqref{eq:proj-wc} predicts in accordance with visual perception the appearance of a counter-phase grating in the central grey bar, see Figures \ref{fig:rec3D} and \ref{fig:rec3D2}. The same result is obtained by the ODOG model, see Figures \ref{fig:recODOG} and \ref{fig:recODOG2}.  In particular, Figures \ref{fig:line_profiles_barrette} and  \ref{fig:line_profiles_barrette2} show higher intensity profile when the background gratings are orthogonal to the central line, while the effect diminishes if the angle of the background decrease from $\pi/2$ to $\pi/3$, see orange and green dashed line. On the other hand, the BIWaM model and the WC-type 2D model for contrast enhancement do not appear suitable to describe this phenomenon.
See for comparison the red and blue dashed lines in Figures \ref{fig:line_profiles_barrette} and  \ref{fig:line_profiles_barrette2}.

\subsection{Poggendorff illusion}  \label{sec:poggendorf}

\begin{figure}[t]
    \centering
    \begin{subfigure}{0.45\textwidth}
    \centering
    \includegraphics[height = 4cm]{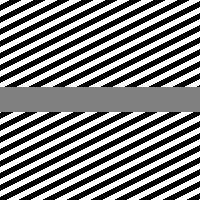}
    \caption{Original image, a variation of the Poggendorff illusion. The presence of the grey central surface induces a misalignment of the background lines.}
    \label{fig:poggendorff-var}
    \end{subfigure}\hfill
    \begin{subfigure}{0.45\textwidth}
       \centering
       \includegraphics[height = 4cm]{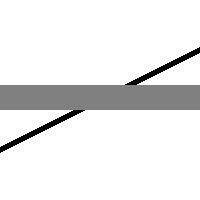} 
   \caption{The classical Poggendorff illusion, extracted from the previous image. The grey surface is superposed on top of the black line, creating a misalignment effect.}
   \label{fig:poggendorff-orig}
   \end{subfigure}
   \\
   \begin{subfigure}{0.45\textwidth}
        \centering
        \includegraphics[height = 4cm]{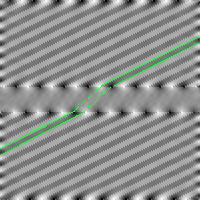}
    \caption{Model output.}
    \label{fig:poggendorff-out}
    \end{subfigure}
    \hfill
    \begin{subfigure}{0.45\textwidth}
    \centering
       \includegraphics[height = 4cm]{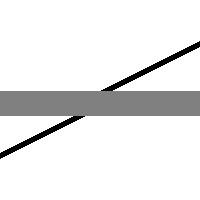}
       \caption{The perceived alignment is reconstructed and isolated after processing.}
       \label{fig:poggendorff-reconstr}
       \end{subfigure}
    \caption{Poggendorff illusion. Parameters: $\sigma_{\mu} = 3$, $\sigma_\omega = 10$, $\lambda = 0.5$.}
    \label{fig:poggendorff}
\end{figure}

The Poggendorff illusion (see Figure~\ref{fig:poggendorff-orig}) consists in the super-position of a surface on top of a continuous line, which then induces a misalignment effect. 
This phenomenon has been deeply investigated \cite{Weintraub1971,Westheimer2008} and studied via neuro-physical experiments, see, e.g., \cite{Weintraub1971}.
Here, we consider a variation of the Poggendorff illusion, where the background is constituted by a grating pattern, see Figure~\ref{fig:poggendorff-var}. Figure \ref{fig:poggendorff-orig} contains the classical Poggendorff illusion, extracted from figure \ref{fig:poggendorff-var}.

\paragraph{Discussion on computational results.} The result obtained by applying the model \eqref{eq:proj-wc} to Figure \ref{fig:poggendorff-var} is presented in Figure~\ref{fig:poggendorff-out}. Similarly as in the previous example, we observe an alternation of oblique bands of different grey intensities within the central grey surface. However, the question here is whether it is possible to reconstruct numerically the perceived misalignment between a fixed black stripe in the bottom part of Figure~\ref{fig:poggendorff-var} and its collinear prosecution lying in the upper part. Note that the perceived alignment differs from the actual geometrical one: for a fixed black stripe in the bottom part, one would in fact perceive the alignment of the corresponding collinear top stripe slightly flushed left, as it is clear from  Figure~\ref{fig:poggendorff-orig} where single stripes have been isolated for better visualisation.
To answer this question, let us look at the reconstruction provided in Figure~\ref{fig:poggendorff-out} and mark by a continuous green line a fixed black stripe in the bottom part of the image. In order to find the stripe in the upper part which is perceived to be collinear with the marked one, one would need to follow how the model propagates the marked stripe across the central surface. By drawing a dashed line in correspondence with such propagation, we can thus find the stripe in the upper part of the image corresponding to the marked one and observe that, as expected, this does not correspond to its actual collinear prosecution. This can be clearly seen in Figure~\ref{fig:poggendorff-reconstr} where the two stripes have been isolated for better visualisation. 

\begin{figure}[t]
\centering
\begin{subfigure}{0.24\textwidth}
    \centering
   \includegraphics[width=\textwidth]{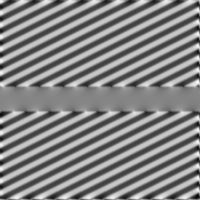}
   \caption{ODOG result.}
   \label{fig:POGG_ODOG}
\end{subfigure}
\begin{subfigure}{0.24\textwidth}
\centering
    \includegraphics[width=\textwidth]{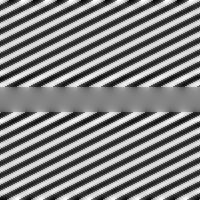}
    \caption{BIWaM result.}
    \label{fig:POGG_BIWaM}
 \end{subfigure}
 \begin{subfigure}{0.24\textwidth}
\centering
    \includegraphics[width=\textwidth]{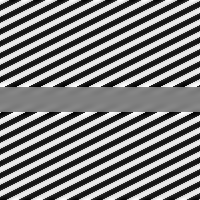}
    \caption{2D WC-type model.}
    \label{fig:POGG_2D}
 \end{subfigure}
 \begin{subfigure}{0.24\textwidth}
\centering
    \includegraphics[width=\textwidth]{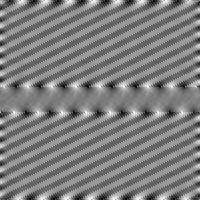}
    \caption{3D WC-type model.}
    \label{fig:POGG_3D}
 \end{subfigure}
    \caption{Reconstruction of the Poggendorff illusion \ref{fig:poggendorff-var} via reference models.  }
    \label{fig:POGG_REF}

\centering
\begin{subfigure}{0.45\textwidth}
    \centering
   \includegraphics[height=4cm]{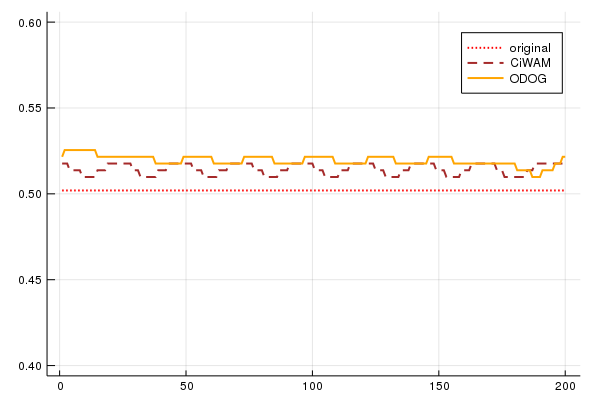}
   \caption{ODOG and BIWaM.}
   \label{fig:lineODOGBIWaM2}
\end{subfigure}
\begin{subfigure}{0.45\textwidth}
\centering
    \includegraphics[height=4cm]{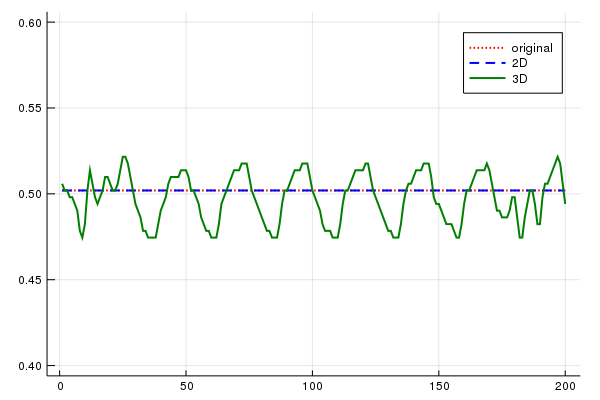}
    \caption{2D VS 3D algorithm.}
    \label{fig:line2D3D2}
 \end{subfigure}
    \caption{Middle line-profiles of outputs in Figure~\ref{fig:POGG_REF}.  }
    \label{fig:POGG-profiles}
\end{figure}

This example shows that the proposed algorithm computes an output in agreement with our perception.  Comparisons with reference models are presented in Figures~\ref{fig:POGG_REF} and \ref{fig:POGG-profiles}.
 We observe that the results obtained via the proposed 3D-WC model cannot be reproduced by the BIWaM nor the WC-type 2D model, which moreover induce non-counter-phase grating in the central grey bar. On the other hand, the result obtained by the ODOG model is consistent with ours, but presents a much less evident alternating grating within the central grey bar. In particular, the induced oblique bands are not visibly connected throughout the whole grey bar, i.e. their induced contrast is very poor and, consequently, the induced edges are not as sharp as the ones reconstructed via our model, see Figure~\ref{fig:POGG-profiles} for one example on the middle-line profile.

\section{Conclusions}

We presented a cortical-inspired setting extending the approach used in \cite{Bertalmio2007,BertalmioFrontiers2014} to describe contrast phenomena in images. By mimicking the structure of V1, the model explicitly takes into account information on  local image orientation and it relates naturally to Wilson-Cowan-type equations introduced in \cite{WilsonCowan1973} to study the evolution of neurons in V1. The model can be efficiently implemented via convolution with appropriate kernels and discretised via standard explicit schemes. The additional information on the local orientation allows to describe contrast/assimilation phenomena as well as notable orientation-dependent illusions outperforming the models introduced in \cite{Bertalmio2007,BertalmioFrontiers2014}.
Furthermore, the performance of the introduced mean field model is competitive with the results obtained by applying some popular orientation-dependent filtering such as the ODOG and the BIWaM models  \cite{Blakeslee1999,Otazu2008}.

Further investigations should address a more accurate modelling reflecting the actual structure of V1. In particular, this concerns the lift operation where the cake wavelet filters should be replaced by Gabor filtering, as well as the interaction weight $\omega$ which could be taken to be the anisotropic heat kernel of \cite{Citti2006} instead of the isotropic Gaussian currently employed.
Finally, extensive numerical experiments should be performed to assess the compatibility of the model with psycho-physical tests measuring the perceptual bias induced by these and other phenomena such as the tilt illusion \cite{Self2014}. This would provide insights about the robustness of the model in reproducing the visual pathway behaviour.

\bibliographystyle{splncs04}
\bibliography{biblio}
\end{document}